\pdfoutput=1
\documentclass[10pt,twocolumn,letterpaper]{article}

\usepackage[pagenumbers]{cvpr}

\usepackage[T1]{fontenc}
\usepackage{cuted}
\usepackage{caption}
\usepackage{microtype}
\usepackage{multirow}
\usepackage{makecell}
\usepackage{tabularx}
\usepackage{colortbl}
\usepackage{array}
\usepackage{bm}
\usepackage{mathtools}
\usepackage[most]{tcolorbox}
\usepackage{placeins}
\usepackage{stfloats}
\usepackage{enumitem}

\makeatletter
\renewcommand\paragraph{\@startsection{paragraph}{4}{\z@}{0.4em \@plus 0.1em}{-0.7em}{\normalfont\normalsize\bfseries}}
\makeatother

\setlist*[itemize]{itemsep=1.5pt,topsep=2.5pt,leftmargin=1.3em}
\setlist*[enumerate]{itemsep=1.5pt,topsep=2.5pt,leftmargin=1.5em}

\definecolor{oursrow}{HTML}{EAF2FB}
\definecolor{subhead}{HTML}{F5F7F9}
\definecolor{frameborder}{HTML}{94A3B8}
\definecolor{framebg}{HTML}{F8FAFC}
\definecolor{mutedgray}{HTML}{64748B}
\definecolor{accentblue}{HTML}{1E5AA8}

\newcommand{\methodname}{\textsc{CineCoder}\xspace}
\newcommand{\benchname}{\textsc{DramaBench}\xspace}

\tcbset{
  figplaceholder/.style={
    enhanced,
    colback=framebg,
    colframe=frameborder,
    boxrule=0.5pt,
    borderline={0.5pt}{0pt}{frameborder,dashed},
    arc=2pt,
    left=6pt, right=6pt, top=5pt, bottom=5pt,
    fontupper=\footnotesize
  }
}

\newcommand{\calS}{\mathcal{S}}

\newcommand{\calJ}{\mathcal{J}}

\definecolor{cvprblue}{rgb}{0.21,0.49,0.74}
\usepackage[pagebackref,breaklinks,colorlinks,allcolors=cvprblue]{hyperref}
\usepackage[capitalize]{cleveref}
\crefname{section}{Sec.}{Secs.}
\Crefname{section}{Section}{Sections}
\Crefname{table}{Table}{Tables}
\crefname{table}{Tab.}{Tabs.}
\crefname{figure}{Fig.}{Figs.}
\Crefname{figure}{Figure}{Figures}

\title{\methodname: Training Visual Coding Agents via Agentic RL\\for Consistent Multi-Shot Video Generation}

\author{%
  Zexuan Yan$^{1}$ \qquad Yue Ma$^{2}$ \qquad Linfeng Zhang$^{1\dagger}$\\[1pt]
  {\small $^{1}$School of Artificial Intelligence, Shanghai Jiao Tong University}\\[-1pt]
  {\small $^{2}$The Hong Kong University of Science and Technology}%
}

\makeatletter
\def\@maketitle{%
  \newpage
  \null
  \vskip .16in
  \begin{center}
    {\Large \bf \@title \par}
    \vspace*{12pt}{%
      \large
      \lineskip .4em
      \begin{tabular}[t]{c}
        \@author
      \end{tabular}
      \par
    }
    \vskip .1em
    \vspace*{2pt}
  \end{center}
}
\makeatother

\begin{document}
\maketitle

\begin{strip}
\centering
\includegraphics[width=\textwidth]{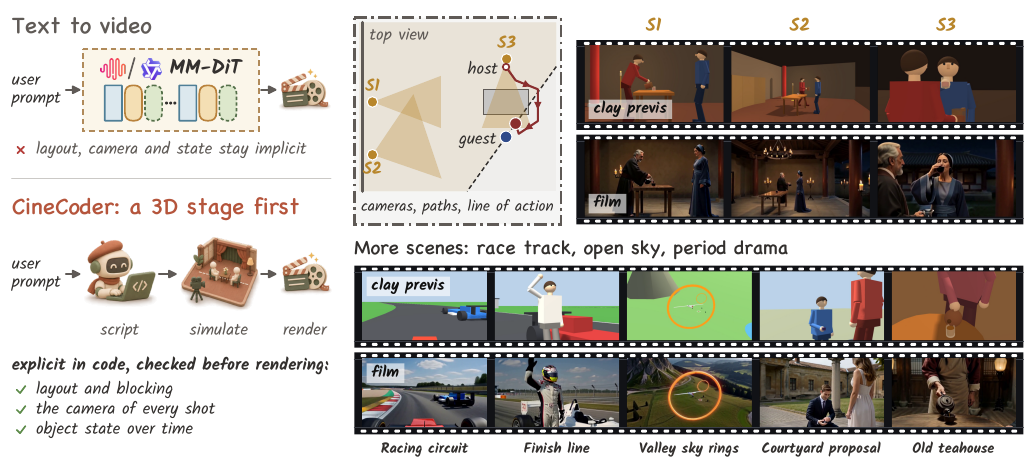}
\captionsetup{hypcap=false}
\captionof{figure}{\textbf{CineCoder: a visual coding framework for multi-shot video synthesis.}
\textbf{Left:} Unlike direct text-to-video generation, \methodname scripts and simulates each multi-shot scene in an executable 3D environment before video synthesis.
\textbf{Top right:} Top-view camera planning maintains consistent spatial layout across cuts and guides the rendered video shot by shot from the 3D preview.
\textbf{Bottom right:} The same pipeline generalizes across diverse settings, including race tracks, open skies, and period dramas.}
\captionsetup{hypcap=true}
\label{fig:teaser}
\end{strip}

\begin{abstract}
Recent advances in video diffusion models and LLM-driven coding agents have enabled promising progress in automated video generation. However, generating coherent multi-shot narrative videos remains challenging: end-to-end diffusion models struggle to preserve object states across cuts and often alter the entire video when a single shot is edited, while existing 3D agents suffer from fragile spatial control and slow environment interaction that bottlenecks interactive coding. To provide a stable world state across cuts, we present \methodname, a visual coding framework that anchors scenes, stateful objects, and characters in executable code before draft-aligned reference-to-video synthesis. When using draft videos as references, it utilize existing frozen DiT models to generate videos across multiple shots while maintaining consistency in draft scenes and motion. Furthermore, to overcome the slow feedback loop that limits reinforcement learning updates in existing agent environments, we develop \textbf{\methodname-Env} in Three.js, a versatile agentic environment that combines diverse scenes, character actions, stateful assets, and camera controls with fast execution and visibility feedback. Building on \methodname-Env, we post-train a 27B multimodal coding agent on real short dramas and synthesized multi-shot scripts, via supervised fine-tuning (SFT) without reasoning traces, followed by Group Relative Policy Optimization (GRPO) with chain-of-thought reasoning to strengthen 3D spatial planning. Finally, we construct \benchname, a comprehensive multi-shot benchmark spanning diverse scene domains, task types, and difficulty levels, where \methodname achieves promising performance across narrative faithfulness, visual quality, and local editing consistency. 
\end{abstract}

\newpage
\section{Introduction}
\label{sec:introduction}

A coherent multi-shot narrative video is governed by a continuous chain of physical and cinematic constraints across cuts. Consider a four-shot dramatic sequence: a wide establishing shot introduces two characters seated across a table; an over-the-shoulder shot shows one character pouring tea into an empty cup; a reverse-angle shot shows the second character opening a box while the filled cup remains on the table; and a close-up captures the final reaction. Directing such a sequence is fundamentally a stateful 3D planning problem rather than merely a frame synthesis task. Each shot must frame its target subjects and actions without occlusion, object and posture changes must persist across cuts, and modifying a single shot---such as changing the final close-up to a high-angle wide view---should leave all preceding shots untouched.

As illustrated in \cref{fig:teaser}, current multi-shot video diffusion models~\citep{meng2025holocine,wang2025multishotmaster,luo2026shotstream} model cross-shot relationships entirely within a shared 2D latent space through windowed cross-attention or causal KV caches. While cross-shot attention helps preserve character appearance and global style, 2D latent attention is not an explicit 3D world model, leading to three systematic failure modes. First, when prompted with a multi-shot story, latent diffusion models struggle to reconcile sharp shot transitions with complex multi-step actions, either merging shots into a single take or sacrificing motion dynamics into near-static frames to preserve cut timing. Second, without persistent 3D state tracking, modified objects frequently revert to their initial states after a cut, so a filled cup becomes empty again or a fallen character stands back up in the next shot. Third, because all shots are entangled in a shared latent trajectory or KV cache, editing the camera angle of a single shot propagates visual drift throughout the video, altering viewpoints and character appearances in earlier unedited shots.

To introduce explicit geometric structure, recent methods construct a 3D scene preview first and render it with a conditional video generator~\citep{zou2026miseenvideo,cozyclay2026,wang2026videocoder}. However, existing 3D agents face major bottlenecks in both spatial representation and interaction speed. Interactive GUI studio agents such as CozyClay~\citep{cozyclay2026} manipulate scenes through dozens of slow tool calls per clip and often retreat from tight over-the-shoulder views to static wide shots to avoid collisions. Meanwhile, coding agents such as Video-Coder~\citep{wang2026videocoder} generate raw Blender scripts without multi-shot camera controls or cross-shot state tracking, where minor numerical errors easily place cameras inside walls or hide key actions behind furniture. Even more critically, to obtain visual rewards during reinforcement learning, Video-Coder must execute offline 3D rendering and full DiT video synthesis before querying an external VLM judge. This minutes-per-rollout pipeline creates a severe feedback bottleneck that makes multi-turn interaction and online reinforcement learning prohibitively slow.

Motivated by these challenges, we propose \methodname, which formulates multi-shot directing as executable visual coding in a fast Three.js environment prior to video synthesis. Our key insight is that reliable multi-shot consistency and scalable agentic learning require pairing a high-level cinematic coding abstraction with sub-second browser-native verification. By anchoring environments, stateful objects, and characters in persistent 3D code, state changes automatically carry over across cuts, and revising a single shot modifies only its local code block while leaving all other shots intact. At the core of our system is \methodname-Env, a Three.js agentic environment where the model composes scenes, stateful assets, character actions, and camera controls without guessing raw 3D coordinates. Crucially, instead of waiting for full DiT video rendering to evaluate a scene, \methodname-Env renders on a headless WebGL2 canvas and inspects frustum rays and canvas buffers directly, returning 3D draft videos, occlusion checks, and multimodal diagnostics in roughly $0.5$\,s per turn.

Building on \methodname-Env, we post-train a 27B multimodal coding agent on multi-shot tasks constructed from real short dramas and synthesized scripts. Because supervising on heterogeneous teacher chain-of-thought traces causes the student to imitate verbose post-hoc text summaries rather than 3D spatial logic, we first apply supervised fine-tuning without teacher reasoning traces so the model masters executable scene syntax and multi-turn repair while preserving its native reasoning prior. We then apply Group Relative Policy Optimization with chain-of-thought reasoning enabled under single-turn rollout pressure, using fast simulator rewards from \methodname-Env to strengthen 3D spatial planning before code generation. Finally, draft-aligned video prompting aligns camera framing, visible actions, and scene appearance descriptions with the rendered 3D draft to prevent text-geometry conflicts during video synthesis.

In summary, our main contributions are three-fold:
\begin{itemize}[itemsep=2.5pt,topsep=3.5pt,parsep=1pt]
  \item \textbf{Visual Coding for Multi-Shot Consistency:} We anchor environments, stateful objects, and characters through executable 3D visual coding, ensuring cross-shot causal consistency and preserving unedited shots during local single-shot editing.
  \item \textbf{\methodname-Env Agentic Environment:} We develop \methodname-Env to support flexible camera movements and rich combinations of scenes, actions, and stateful assets, providing fast draft generation and multimodal visibility feedback for online reinforcement learning.
  \item \textbf{Post-Trained 27B Multimodal Coding Agent:} We post-train a 27B multimodal coding agent via two-stage supervised fine-tuning and reinforcement learning with draft-aligned prompting, demonstrating substantial gains across extensive evaluations on \benchname.
\end{itemize}

\newpage
\section{Related Work}
\label{sec:related_work}

\begin{table}[!t]
\centering
\caption{\textbf{Comparison with prior multishot methods.}}
\label{tab:related}
\setlength{\tabcolsep}{7.1pt}
\renewcommand{\arraystretch}{0.96}
\footnotesize
\resizebox{\columnwidth}{!}{%
\begin{tabular}{l cccc}
\toprule
\textbf{Method} & \makecell{\textbf{World}\\\textbf{State}} & \makecell{\textbf{Multi}\\\textbf{Shot}} & \makecell{\textbf{Local}\\\textbf{Edit}} & \makecell{\textbf{Visual}\\\textbf{RL}} \\
\midrule
StoryDiffusion~\citep{zhou2024storydiffusion} & \multirow{2}{*}{2D Keyframe} & \multirow{2}{*}{\checkmark} & \multirow{2}{*}{$\times$} & \multirow{2}{*}{$\times$} \\
MovieDreamer~\citep{zhao2024moviedreamer} & & & & \\
\arrayrulecolor{black!20}\specialrule{0.25pt}{1.0pt}{1.0pt}\arrayrulecolor{black}
HoloCine~\citep{meng2025holocine} & \multirow{3}{*}{Latent Attn} & \multirow{3}{*}{\checkmark} & \multirow{3}{*}{$\times$} & \multirow{3}{*}{$\times$} \\
MultiShotMaster~\citep{wang2025multishotmaster} & & & & \\
ShotStream~\citep{luo2026shotstream} & & & & \\
\midrule
Code2Video~\citep{chen2026code2video} & 2D Code & \checkmark & \checkmark & $\times$ \\
JanusCoder~\citep{sun2026januscoder} & 2D Code & $\times$ & \checkmark & $\times$ \\
\arrayrulecolor{black!20}\specialrule{0.25pt}{1.0pt}{1.0pt}\arrayrulecolor{black}
CozyClay~\citep{cozyclay2026} & 3D Previs & \checkmark & \checkmark & $\times$ \\
StateFlow~\citep{yin2026stateflow} & 3D State & \checkmark & \checkmark & $\times$ \\
\arrayrulecolor{black!20}\specialrule{0.25pt}{1.0pt}{1.0pt}\arrayrulecolor{black}
Video-Coder~\citep{wang2026videocoder} & 3D Code & $\times$ & \checkmark & \checkmark \\
\rowcolor{oursrow}\textbf{\methodname} & \textbf{3D Code} & \textbf{\checkmark} & \textbf{\checkmark} & \textbf{\checkmark} \\
\bottomrule
\end{tabular}%
}
\end{table}

\begin{figure*}[!t]
\centering
\includegraphics[width=0.92\textwidth]{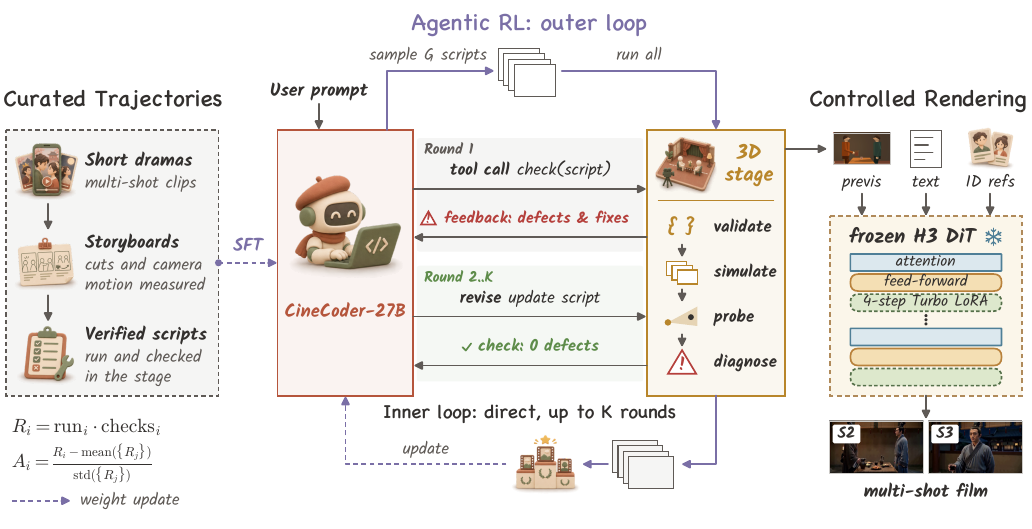}
\caption{\textbf{Overview of \methodname.}
\textbf{Curated Trajectories:} Multi-shot real dramas are converted into verified 3D scene scripts for supervised fine-tuning.
\textbf{Inner Loop:} Given a story prompt, the agent executes code in the 3D environment and iteratively revises its script using diagnostic feedback.
\textbf{Agentic RL:} Across parallel rollouts, the agent samples $G$ trajectories, evaluates them in the 3D environment, and updates its policy via GRPO using simulator rewards.
\textbf{Controlled Rendering:} The 3D preview video, user prompt, and identity reference images are fed into DiT model to generate multi-shot video.}
\label{fig:overview}
\end{figure*}

\paragraph{Multi-Shot Video Generation.}
Diffusion transformers~\citep{yang2024cogvideox,kong2024hunyuanvideo,wan2025wan} have been extended from single clips to multi-shot generation along two main directions. Keyframe-based pipelines~\citep{zhou2024storydiffusion,zhao2024moviedreamer} generate storyboards first and animate each shot separately, which preserves rough character appearance but loses 3D scene consistency across cuts. Joint and streaming multi-shot models including HoloCine~\citep{meng2025holocine}, MultiShotMaster~\citep{wang2025multishotmaster}, and ShotStream~\citep{luo2026shotstream} connect shots through windowed cross-attention or causal KV caches, while camera-conditioned models~\citep{wang2023motionctrl,he2024cameractrl,yang2026shotverse} inject camera trajectories into video diffusion. However, because latent attention lacks an explicit 3D world state, these models struggle to balance accurate shot cuts with dynamic motion, modified objects often revert to their initial states after a cut, and editing a single shot alters the entire video. \methodname addresses these issues by representing multi-shot world states and camera trajectories in executable Three.js code prior to video synthesis.

\paragraph{Visual Coding Agents and 3D Previsualization.}
An alternative to purely latent generation is to represent scenes programmatically with language models, including 3D scene and layout generation~\citep{yang2024holodeck,hu2024scenecraft,sun20253dgpt}, 2D visual programs and animations~\citep{chen2026code2video,sun2026januscoder}, camera trajectory planning~\citep{wu2026campilot,zhang2026filmflow}, multi-shot agentic self-evolution~\citep{zhang2026cineevolve}, and interactive 3D previsualization~\citep{zou2026miseenvideo,yin2026stateflow,cozyclay2026}. However, GUI-based studio agents such as CozyClay~\citep{cozyclay2026} depend on slow multi-turn tool calls and often fall back to distant wide shots to avoid collisions. Meanwhile, single-shot Blender coding agents~\citep{li2026videococo,wang2026videocoder} and code-based world-modeling systems~\citep{fang2026code} focus on single-clip physics or open-ended world rollouts rather than multi-shot narrative directing with cross-cut camera grammar. Closest to our training setup, Video-Coder~\citep{wang2026videocoder} trains a single-shot 3D coding model with reinforcement learning, but relies on full DiT video rendering and an external VLM judge to obtain visual rewards. As summarized in \cref{tab:related}, \methodname extends visual coding agents to multi-shot directing in \methodname-Env with persistent world state and raycast visibility checks, trains spatial reasoning via fast simulator rewards without DiT inference latency, and aligns video prompts with the rendered 3D scene.

\newpage
\section{Method}
\label{sec:method}

As shown in \cref{fig:overview}, \methodname decomposes multi-shot video generation into three parts: coding, simulating and rendering. 
We propose \textbf{\methodname-Env} for coding, \textbf{Agentic RL} in simulating, and \textbf{V2V-DiT} for controlled rendering. First, \cref{sec:method_stage} introduces \textbf{\methodname-Env}, a Three.js environment that converts high-level scene scripts into persistent 3D trajectories and verifies on-screen visibility. Building on this environment, \cref{sec:method_rl} describes how we construct multi-turn training trajectories from real short dramas and train our coding agent via SFT followed by RL. Finally, \cref{sec:method_handoff} explains how draft-aligned prompt generation converts the resulting 3D draft videos into photorealistic videos without scene-drifting.

\subsection{The \methodname-Env 3D Environment}
\label{sec:method_stage}

\begin{figure}[!t]
\centering
\includegraphics[width=\columnwidth]{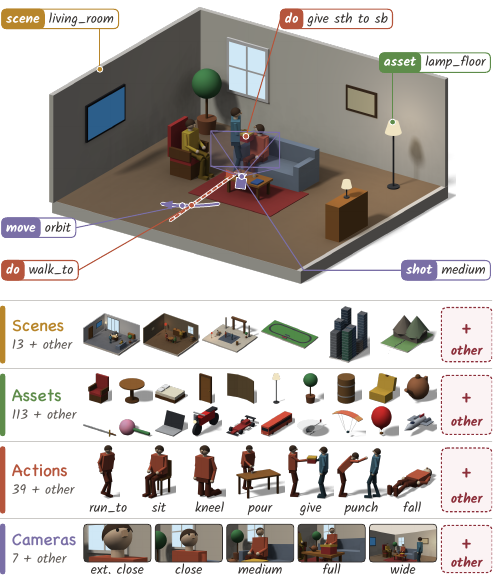}
\caption{\textbf{The \methodname 3D environment.} A multi-shot scene is specified through four high-level components: scenes, objects, actions, and cameras.}
\label{fig:harness}
\end{figure}

\paragraph{High-Level Directing with Available Assets.}
When an LLM writes raw Three.js or Blender code from scratch, most of its capacity is spent guessing 3D coordinates, rotations, and camera matrices, where small numerical errors easily push characters out of frame or behind walls. \methodname-Env separates high-level scene directing from low-level 3D geometry calculations. As shown in \cref{fig:harness}, a script specifies a multi-shot video through four modular components: preset indoor and outdoor environments or custom room dimensions, stateful objects and vehicles, character and vehicle actions, and standard film camera controls covering shot scale, viewing angle, camera height, and camera motion. Given the scripted actions and camera choices, the environment automatically computes collision-free character paths and 3D camera trajectories across cuts.

\paragraph{Persistent World State and Visibility Verification.}
Beyond simplifying layout within a single shot, \methodname-Env maintains a continuous 3D world state $\calS_\tau$ across cuts over simulation frames $\tau$, tracking character postures, held objects, liquid levels in containers, open or closed lids, and ongoing vehicle or falling motion. Any object picked up, filled, or opened in Shot~$k$ remains in that exact state in Shot~$k+1$, and state transitions occur only when physical preconditions such as arm reachability are satisfied. Crucially, executing an action in 3D space does not guarantee that it is visible in the camera view. In the inner loop of \cref{fig:overview}, the environment checks code syntax, simulates the scene to record all frame trajectories, and casts rays from the camera frustum to verify that target characters and key actions remain unobstructed on screen and that adjacent shots maintain smooth camera transitions. Whenever a physical precondition or visibility check fails, \methodname-Env returns a structured diagnostic message and clay preview so the agent can revise its script efficiently.

\subsection{Data Pipeline and Agentic RL}
\label{sec:method_rl}

\paragraph{Constructing Tasks for Supervised Fine-Tuning(SFT).}
While \methodname-Env provides a fast, verifiable interface, training a coding agent for multi-shot directing requires paired multi-shot stories and Three.js programs, which is rare in existing datasets. We therefore build a training corpus with strictly non-overlapping train and test drama series. We combine synthesized multi-shot scripts with real short-drama episodes parsed via shot-boundary detection and VLM captioning into multi-shot story prompts. Teacher models interact with \methodname-Env for up to $K$ repair turns; verified scripts are used for SFT, while harder prompts are reserved for RL. Because different teacher models produce inconsistent or post-hoc summarized chain-of-thought traces, supervising on teacher reasoning causes the student model to imitate surface text patterns rather than 3D spatial logic. We therefore perform SFT without chain-of-thought traces by removing teacher reasoning and supervising only on the final verified code in each trajectory, preserving the base model's native reasoning capability for RL.

\begin{figure*}[!t]
\centering
\includegraphics[width=\textwidth]{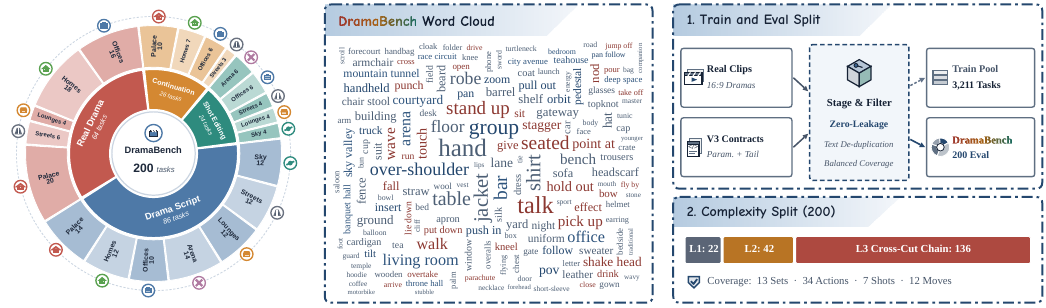}
\caption{\textbf{Overview of the $200$-task \benchname benchmark.}
\textbf{Left:} Task distribution across four evaluation types on the inner ring and seven scene categories on the outer ring.
\textbf{Center:} Common scene, camera, and action terms in the benchmark prompts.
\textbf{Right:} Data construction pipeline with non-overlapping train and test splits at the top, followed by task complexity levels and action coverage below.}
\label{fig:dramabench}
\end{figure*}

\paragraph{Bootstrapping Spatial Planning via RL.}
{%
\setlength{\abovedisplayskip}{3.5pt plus 1pt minus 1pt}%
\setlength{\belowdisplayskip}{3.5pt plus 1pt minus 1pt}%
\setlength{\abovedisplayshortskip}{2pt plus 1pt}%
\setlength{\belowdisplayshortskip}{2pt plus 1pt}%
Although SFT teaches code syntax and multi-turn repair, complex multi-shot prompts require planning character positions, camera angles, and cross-shot visibility before writing code. We therefore further train the policy using GRPO~\citep{shao2024deepseekmath,guo2025deepseekr1} driven by fast simulator feedback from \methodname-Env.

As shown in the outer loop of \cref{fig:overview}, for each input prompt $x$, the policy samples a group of $G$ candidate trajectories $\{o_1, \dots, o_G\} \sim \pi_{\theta_{\mathrm{old}}}(\cdot \mid x)$ and executes them in \methodname-Env. Each rollout $i$ receives a reward $R_i \in [0, 1]$ defined as a function of two verification scores:
\begin{equation}
  R_i = \mathrm{run}_i \cdot f_{\mathrm{stage}}\!\big(S_i^{\mathrm{sim}},\, S_i^{\mathrm{con}}\big),
  \label{eq:grpo_reward}
\end{equation}
where $\mathrm{run}_i \in \{0, 1\}$ indicates whether the script executes without error, $S_i^{\mathrm{sim}}$ measures scene verification quality across visibility and camera continuity checks, and $S_i^{\mathrm{con}}$ measures the fraction of satisfied story checklist items. To evaluate candidate rollouts for the same prompt without a separate critic network, we normalize the rewards within each group to obtain the relative advantage:
\begin{equation}
  A_i = \frac{R_i - \mathrm{mean}(\{R_j\}_{j=1}^G)}{\mathrm{std}(\{R_j\}_{j=1}^G) + \varepsilon}.
  \label{eq:grpo_adv}
\end{equation}
We then assign $A_i$ to the tokens of trajectory $o_i$ and update the policy $\pi_\theta$ by maximizing a KL-regularized clipped objective with respect to the SFT reference policy $\pi_{\mathrm{ref}}$:
\begin{equation}
  \calJ_{\mathrm{GRPO}}(\theta) = \mathbb{E}\!\Big[\frac{1}{G}\sum_{i=1}^{G} \mathrm{clip}(\rho_i)\, A_i - \beta\, \mathbb{D}_{\mathrm{KL}}(\pi_\theta \,\|\, \pi_{\mathrm{ref}})\Big],
  \label{eq:grpo_obj}
\end{equation}
where $\rho_i$ is the importance sampling ratio between $\pi_\theta$ and $\pi_{\theta_{\mathrm{old}}}$. Crucially, how rollouts are sampled determines what skill RL improves. When rollouts are sampled without chain-of-thought reasoning across multiple repair turns, the model tends to rely on simulator error messages rather than improving its initial scene plan. In our RL stage, we enable native chain-of-thought reasoning during rollouts, limit training rollouts to a single draft turn so the model must plan the scene accurately on the first attempt, and train on a deduplicated prompt pool balanced with real short dramas. Because $A_i$ updates both the reasoning tokens and the generated code, GRPO directly strengthens 3D spatial planning before code generation.%
}

\subsection{Draft-Aligned Video Rendering}
\label{sec:method_handoff}

Given the simulated 3D world state $\calS_\tau$, our rendering stage translates the 3D draft into the final multi-shot video via draft-aligned prompting:
\begin{equation}
  V = \mathrm{DiT}_{\mathrm{Ref2V}}\!\big(V_{\mathrm{3D}},\, I_{\mathrm{ref}},\, p(\calS_\tau)\big),
  \label{eq:draft_rendering}
\end{equation}
where $V$ is the final output video, $\mathrm{DiT}_{\mathrm{Ref2V}}$ is the frozen four-step H3 Ref2VA diffusion model, $V_{\mathrm{3D}}$ is the rendered 3D draft video fixing cuts and trajectories, $I_{\mathrm{ref}}$ denotes the character reference portraits, and $p(\calS_\tau)$ is the draft-aligned text prompt derived from $\calS_\tau$. Specifically, for each shot, camera descriptions in $p(\calS_\tau)$ are measured from the actual on-screen framing and character screen positions rather than copied from the input script; action descriptions retain only events verified as visible on screen by raycasting; and our 27B model generates detailed descriptions of the visible room layout, lighting, and surface materials while omitting occluded objects, preventing text-geometry override and untextured graybox bleed-through (\cref{fig:overview}, right).

\section{Experiments}
\label{sec:experiments}

We evaluate \methodname across three core aspects: how our visual coding paradigm compares against end-to-end multi-shot diffusion and prior 3D coding methods on narrative faithfulness and local shot editing in \cref{sec:exp_main}; how multi-turn diagnostic feedback and our two-stage training pipeline improve both 3D scene scripting and output video quality in \cref{sec:exp_ablation}; and how draft-aligned video prompting improves camera alignment and visual realism in \cref{sec:exp_handoff}.

\subsection{Experimental Setup}
\label{sec:exp_setup}

\begin{table*}[!t]
\centering
\caption{\textbf{Quantitative comparison on multi-shot video generation.}
We compare end-to-end multi-shot diffusion models and 3D-guided video generation methods on shot cut accuracy, prompt success rate, causal VQA accuracy, character consistency, visual quality, and local editing consistency. Best results are in \textbf{bold} and second best are \underline{underlined}.}
\label{tab:main}
\setlength{\tabcolsep}{6.2pt}
\renewcommand{\arraystretch}{1.02}
\footnotesize
\resizebox{\textwidth}{!}{%
\begin{tabular}{ll ccccc cc}
\toprule
\textbf{Category} & \textbf{Method} & \textbf{SCA}$\uparrow$ & \textbf{Prompt-SR\%}$\uparrow$ & \textbf{Causal-VQA\%}$\uparrow$ & \textbf{ID-Sim}$\uparrow$ & \textbf{VBench}$\uparrow$ & \textbf{Edit-PSNR}$\uparrow$ & \textbf{Edit-DINOv2}$\uparrow$ \\
\midrule
\multirow{5}{*}{Multi-Shot DiT} & HoloCine~\citep{meng2025holocine} & 0.901 & 79.2 & \textbf{99.0} & 0.862 & 0.802 & 14.35 & 0.660 \\
 & MultiShotMaster~\citep{wang2025multishotmaster} & \underline{0.968} & 70.8 & 93.9 & 0.875 & 0.776 & 12.97 & 0.553 \\
 & ShotStream~\citep{luo2026shotstream} & \textbf{0.981} & 50.0 & 80.8 & \textbf{0.969} & 0.705 & 12.22 & 0.698 \\
 & Seedance 2.5 & 0.722 & 62.5 & 93.8 & 0.851 & 0.802 & 14.40 & 0.360 \\
 & MiniMax-H3 Direct & 0.500 & 8.3 & 86.5 & 0.781 & 0.791 & 12.30 & 0.672 \\
\midrule
 & CozyClay~\citep{cozyclay2026} & 0.897 & 62.5 & 91.2 & \underline{0.917} & 0.772 & \textbf{17.86} & \textbf{0.882} \\
 & Video-Coder~\citep{wang2026videocoder} & 0.563 & 25.0 & 68.4 & 0.840 & 0.798 & 13.35 & 0.597 \\
\rowcolor{oursrow}\cellcolor{white} & \textbf{\methodname-SFT} & 0.927 & \underline{83.3} & 95.0 & 0.909 & \underline{0.811} & 17.30 & \underline{0.859} \\
\rowcolor{oursrow}\cellcolor{white}\multirow{-4}{*}{3D Previs + V2V} & \textbf{\methodname-RL} & 0.942 & \textbf{95.8} & \underline{98.3} & \underline{0.917} & \textbf{0.819} & \underline{17.85} & 0.836 \\
\bottomrule
\end{tabular}%
}
\end{table*}

\begin{table*}[!t]
\centering
\caption{\textbf{Ablation of agent training stages on \benchname video generation.}
All rows share the video rendering pipeline of \cref{tab:main} and vary only the model checkpoint, showing that improved 3D scene planning directly translates into higher prompt success and causal accuracy in the output videos. Best results are in \textbf{bold} and second best are \underline{underlined}.}
\label{tab:ablation_video}
\setlength{\tabcolsep}{11.3pt}
\renewcommand{\arraystretch}{1.02}
\footnotesize
\resizebox{\textwidth}{!}{%
\begin{tabular}{l ccccc cc}
\toprule
\textbf{Training Variant} & \textbf{SCA}$\uparrow$ & \textbf{Prompt-SR\%}$\uparrow$ & \textbf{Causal-VQA\%}$\uparrow$ & \textbf{ID-Sim}$\uparrow$ & \textbf{VBench}$\uparrow$ & \textbf{Edit-PSNR}$\uparrow$ & \textbf{Edit-DINOv2}$\uparrow$ \\
\midrule
Base-27B & 0.903 & 60.9 & 93.6 & 0.906 & 0.803 & 17.70 & \underline{0.843} \\
+ SFT  & \underline{0.927} & 83.3 & 95.0 & 0.909 & \underline{0.811} & 17.30 & \textbf{0.859} \\
\quad + RL w/o CoT & 0.919 & \underline{87.5} & \underline{97.1} & \underline{0.912} & 0.810 & \underline{17.83} & 0.809 \\
\rowcolor{oursrow}\quad + RL w/ CoT & \textbf{0.942} & \textbf{95.8} & \textbf{98.3} & \textbf{0.917} & \textbf{0.819} & \textbf{17.85} & 0.836 \\
\bottomrule
\end{tabular}%
}
\end{table*}

\paragraph{Benchmark and Implementation Details.}
Existing video benchmarks focus on clips and lack ground-truth verification for cross-shot object states. We therefore construct \benchname, a $200$-task multi-shot benchmark illustrated in \cref{fig:dramabench}. With all test drama series held out and filtered via  deduplication, \benchname spans four complementary task categories: \emph{Drama Script} for multi-step object interactions, \emph{Real Drama} extracted from unseen short-drama episodes from streaming platforms, \emph{Continuation} requiring cross-scene state memory, and \emph{Shot Editing} for localized camera changes. We initialize our agent from Qwen3.8-27B with rank-$64$ LoRA adapters, training Stage-1 SFT without chain-of-thought(COT) traces for $3$ epochs on verified scripts and Stage-2 GRPO with COT reasoning using group size $G=4$. At evaluation, all coding agents can revise their scripts for up to $K=3$ turns using execution feedback; full benchmark details and hyperparameters are provided in the supplement.

\paragraph{Baselines and Evaluation Metrics.}
In \cref{tab:main}, we compare \methodname-SFT and \methodname-RL against two major baseline families using the same frozen four-step H3 Ref2VA renderer for all 3D-guided methods. The \emph{Multi-Shot DiT} group evaluates end-to-end diffusion models that generate multi-shot videos directly from text without explicit 3D state: HoloCine~\citep{meng2025holocine}, MultiShotMaster~\citep{wang2025multishotmaster}, ShotStream~\citep{luo2026shotstream}, Seedance~2.5, and text-only MiniMax-H3 Direct. The \emph{3D Previs + V2V} group evaluates the interactive 3D studio agent CozyClay~\citep{cozyclay2026} and the open-weight Blender RL coding model Video-Coder~\citep{wang2026videocoder}. We assess generated videos across seven metrics covering shot boundary alignment via Shot Cut Accuracy(SCA)~\citep{meng2025holocine}, narrative and state faithfulness via whole-video Prompt Success Rate(Prompt-SR/P-SR) and fine-grained Causal-VQA accuracy, character consistency via DreamSim ID-Sim, overall visual quality across six dimensions of VBench~\citep{huang2024vbench}, and consistency of unedited shots via Edit-PSNR and Edit-DINOv2 when modifying shot's camera parameters under a fixed random seed.

\begin{figure*}[!t]
\centering
\includegraphics[width=\textwidth]{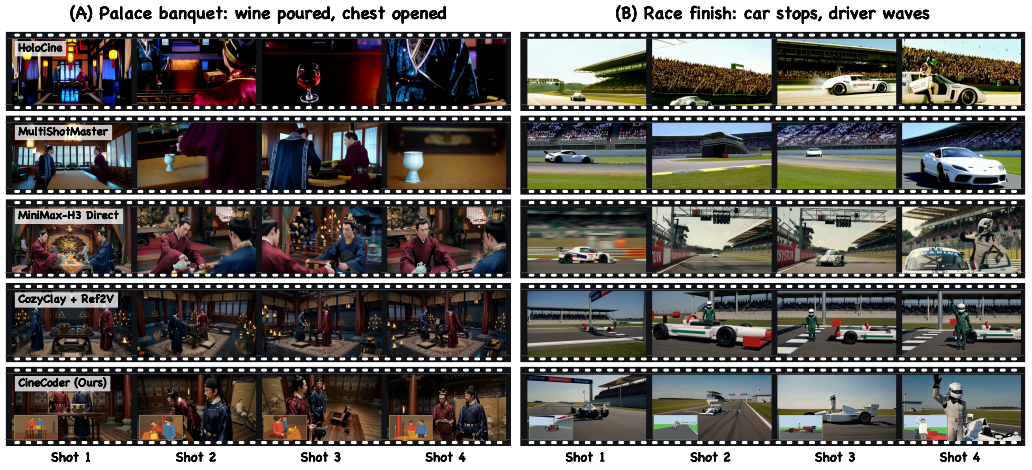}
\caption{\textbf{Qualitative comparison on multi-shot video generation.} Across both an indoor multi-character drama and an outdoor interactive scene, \methodname maintains consistent character identity, accurate camera cuts, and coherent object states compared with existing baselines.}
\label{fig:qual_main}
\end{figure*}

\begin{figure}[!t]
\centering
\includegraphics[width=0.78\columnwidth]{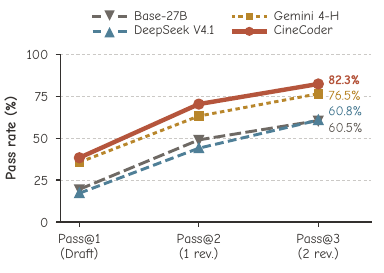}
\caption{\textbf{Effect of multi-turn diagnostic repair.} Task pass rates on \benchname from the initial script draft through two rounds of execution feedback across Base-27B, teacher models, and \methodname.}
\label{fig:repair}
\end{figure}

\begin{figure*}[!t]
\centering
\includegraphics[width=\textwidth]{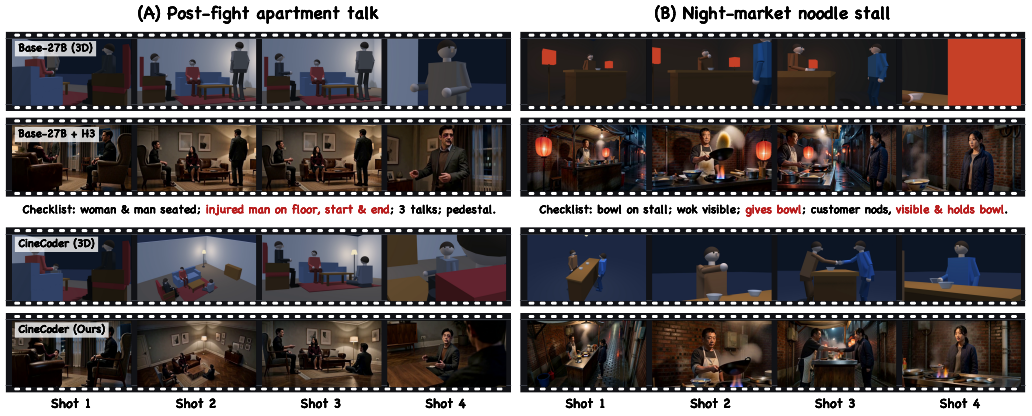}
\caption{\textbf{Qualitative comparison between Base-27B and \methodname{}.} For each example, the 3D scene preview is shown above the final rendered video for a scene-continuation task on the left and an outdoor interaction task on the right, with failed story requirements highlighted in \textcolor[HTML]{B91C1C}{red}.}
\label{fig:qual_ablation}
\end{figure*}

\subsection{Main Results}
\label{sec:exp_main}

\Cref{tab:main} and \cref{fig:qual_main} highlight a clear trade-off between end-to-end latent diffusion and 3D-guided generation. When prompted with a multi-shot story directly, diffusion baselines such as MiniMax-H3 Direct and Seedance~2.5 frequently merge cuts or miss intermediate actions, resulting in low shot-cut accuracy and prompt success. Multi-shot diffusion models conditioned on per-shot captions achieve much sharper cut boundaries, but struggle to maintain dynamic action and causal state simultaneously: ShotStream achieves high cut accuracy largely by generating near-static frames, which degrades its motion quality and prompt success, whereas HoloCine misses strict cut boundaries on complex transitions. More importantly, because all shots are coupled within a shared latent space or causal KV cache, editing the camera angle of the final shot causes severe visual drift across earlier, unedited shots for every Multi-Shot DiT baseline. In particular, because ShotStream conditions subsequent shots on the cached features of the first shot, changing a single shot's viewpoint alters camera angles and character appearance throughout the entire video.

Introducing an explicit 3D preview in \emph{3D Previs + V2V} decouples local shot editing and state tracking from video synthesis, provided that the upstream 3D script is accurate. CozyClay preserves unedited shots well during camera edits mainly because its GUI agent defaults to distant static wide shots to avoid collisions, which limits its ability to capture close-up interactions and reduces narrative success. Conversely, Video-Coder~\citep{wang2026videocoder} generates raw Blender coordinates without multi-shot camera controls or state tracking, frequently collapsing multi-shot stories into a single continuous take and performing poorly across both cut accuracy and narrative faithfulness. By combining \methodname-Env with supervised fine-tuning, \methodname-SFT already surpasses all prior methods on both prompt success and overall visual quality. Further reinforcing spatial planning with RL elevates \methodname-RL across cut accuracy, character consistency, and visual quality, reaching $95.8\%$ Prompt-SR, $98.3\%$ Causal-VQA accuracy, and the best VBench score of $0.819$ while maintaining strong local editing consistency.

\subsection{Ablations on Agentic Framework}
\label{sec:exp_ablation}

\Cref{tab:ablation_video} and \cref{fig:repair,fig:qual_ablation} examine how closed-loop feedback in \methodname-Env and our two-stage training recipe contribute to both upstream visual coding and downstream video generation.

\paragraph{Multi-Turn Diagnostic Repair.}
\Cref{fig:repair} tracks task pass rates on the 200-task \benchname benchmark from the initial draft through two rounds of execution feedback. Because \methodname-Env pinpoints visibility occlusions and camera transition errors directly from simulation, iterative self-repair consistently rescues failed initial drafts across all evaluated models, more than doubling the pass rate of both Base-27B and \methodname over two revision turns and allowing our 27B agent to surpass zero-shot frontier teacher models.

\paragraph{Effect of Two-Stage SFT and RL Training.}
Beyond test-time repair, our two-stage training progressively strengthens the agent's spatial planning ability. Removing noisy teacher reasoning traces during SFT while re-enabling the base model's native reasoning at test time raises the three-turn \benchname pass rate by nearly $20$ percentage points over the base model, whereas disabling test-time reasoning causes a sharp drop, confirming that SFT should teach clean scene syntax without overwriting native spatial reasoning. In the RL stage, running GRPO without COT reasoning across multi-turn rollouts improves multi-turn repair but leaves first-draft accuracy nearly unchanged, as reasoning traces shrink and many rollout groups receive tied rewards. By contrast, enabling COT reasoning under single-turn rollout pressure forces the policy to plan character layout and camera viewpoints before writing code. This substantially reduces tied rollout groups and delivers the strongest first-draft and multi-turn pass rates alongside the cleanest visibility scores; detailed per-category results are reported in the supplement.

\paragraph{Impact of 3D Scene Quality on Output Video.}
\Cref{fig:qual_ablation} pairs 3D scene previews directly above their rendered videos on two representative tasks, demonstrating that downstream video errors originate from upstream 3D staging mistakes rather than diffusion noise. In Scene~A on the left, Base-27B ignores the prior scene state and places the injured character standing throughout the continuation, whereas \methodname preserves the fallen posture across cuts in both the 3D preview and the output video. In Scene~B on the right, Base-27B fails reachability and occlusion checks so the object handover is hidden from view, whereas \methodname stages a clear interaction. \Cref{tab:ablation_video} confirms this trend quantitatively on \benchname under fixed rendering settings: each training stage consistently improves shot-cut alignment, prompt success, causal VQA accuracy, character consistency, and VBench visual quality in the rendered videos, with CoT-enabled RL outperforming RL without CoT across every video metric and driving a $12.5$-point gain in end-to-end prompt success over SFT.

\begin{table}[!t]
\centering
\caption{\textbf{Ablation of 3D-to-video rendering components on \benchname.}
Aligning character descriptions with the rendered camera view and adding scene appearance descriptions improves shot-cut accuracy and visual realism while preserving unedited shots during local edits. \textbf{P-SR} denotes Prompt Success Rate; best results are in \textbf{bold} and second best are \underline{underlined}.}
\label{tab:handoff}
\setlength{\tabcolsep}{3.5pt}
\renewcommand{\arraystretch}{1.14}
\footnotesize
\resizebox{\columnwidth}{!}{%
\begin{tabular}{ccc cccc}
\toprule
\textbf{3D} & \textbf{Character} & \textbf{Scene} & \multirow{2}{*}{\textbf{SCA}$\uparrow$} & \multirow{2}{*}{\textbf{P-SR}$\uparrow$} & \multirow{2}{*}{\textbf{NIQE}$\downarrow$} & \multirow{2}{*}{\textbf{Edit-DINOv2}$\uparrow$} \\[-1.5pt]
\textbf{Preview} & \textbf{Alignment} & \textbf{Descriptions} & & & & \\
\midrule
$\times$ & $\times$ & $\times$ & 0.500 & 8.3 & \underline{4.24} & 0.672 \\
\checkmark & $\times$ & $\times$ & 0.878 & \underline{79.2} & 6.46 & 0.838 \\
\checkmark & \checkmark & $\times$ & \underline{0.923} & \underline{79.2} & 4.94 & \textbf{0.876} \\
\rowcolor{oursrow}\textbf{\checkmark} & \textbf{\checkmark} & \textbf{\checkmark} & \textbf{0.927} & \textbf{83.3} & \textbf{4.05} & \underline{0.859} \\
\bottomrule
\end{tabular}%
}
\end{table}

\subsection{Draft-Aligned Prompting and Local Editing}
\label{sec:exp_handoff}

Finally, \cref{tab:handoff} evaluates the role of each rendering component when translating a fixed 3D preview into final video on \benchname. Generating video from text alone without the 3D preview loses shot timing and action order, resulting in very low prompt success and severe drift during single-shot edits. Providing the continuous 3D preview video immediately establishes spatial layout and boosts narrative success, but pairing it with a naive script prompt still leaves a gap: original camera labels can conflict with the rendered 3D viewpoint, and unmentioned 3D character models or walls remain untextured in the output video. Aligning character descriptions with the measured camera framing resolves the viewpoint conflict and improves both cut accuracy and edit stability, while adding scene appearance descriptions generated by our 27B model renders the entire environment with realistic textures, achieving the best overall shot-cut alignment, prompt success, and visual realism.

\FloatBarrier
\section{Conclusion}
\label{sec:conclusion}

We presented \methodname, an agentic coding framework for multi-shot video synthesis. By directing multi-shot scenes through high-level actions, stateful assets, and camera controls in \methodname-Env, our approach maintains persistent 3D object states and local editability across cuts. Post-training our 27B multimodal coding agent via supervised fine-tuning without teacher reasoning traces followed by reinforcement learning with chain-of-thought reasoning substantially improves both multi-turn error repair and first-draft 3D spatial planning. Combined with draft-aligned video prompting, \methodname delivers promising performance on \benchname.

{
    \small
    \bibliographystyle{ieeenat_fullname}
    \bibliography{references}

@inproceedings{chen2026code2video,
  title     = {{Code2Video: A Code-centric Paradigm for Educational Video Creation}},
  author    = {Yanzhe Chen and Kevin Qinghong Lin and Mike Zheng Shou},
  booktitle = {International Conference on Machine Learning (ICML)},
  volume    = {306},
  series    = {Proceedings of Machine Learning Research},
  pages     = {15370--15396},
  year      = {2026}
}

@misc{cozyclay2026,
  title        = {{CozyClay: Open-Source 3D Previsualization Software for AI Video Workflows}},
  author       = {{CozyClay Contributors}},
  howpublished = {\url{https://github.com/HaD0Yun/CozyClay}},
  note         = {Version 3.2.2},
  year         = {2026}
}

@article{fang2026code,
  title   = {{Code Plans, Diffusion Renders: Open-Ended Generative World Modeling}},
  author  = {Zixun Fang and Yawen Shao and Kai Zhu and Jie Xiao and Shihan Chen and Yu Liu and Xueyang Fu and Yang Cao and Wei Zhai and Zheng-Jun Zha},
  journal = {arXiv preprint arXiv:2609.26458},
  year    = {2026}
}

@article{guo2025deepseekr1,
  title   = {{DeepSeek-R1 incentivizes reasoning in LLMs through reinforcement learning}},
  author  = {Daya Guo and Dejian Yang and Haowei Zhang and Junxiao Song and Peiyi Wang and Qihao Zhu and Runxin Xu and Ruoyu Zhang and Shirong Ma and Xiao Bi and others},
  journal = {Nature},
  volume  = {645},
  number  = {8081},
  pages   = {633--638},
  year    = {2025},
  doi     = {10.1038/s41586-025-09422-z}
}

@inproceedings{he2024cameractrl,
  title     = {{CameraCtrl: Enabling Camera Control for Video Diffusion Models}},
  author    = {Hao He and Yinghao Xu and Yuwei Guo and Gordon Wetzstein and Bo Dai and Hongsheng Li and Ceyuan Yang},
  booktitle = {International Conference on Learning Representations (ICLR)},
  pages     = {100433--100464},
  year      = {2025}
}

@inproceedings{hu2024scenecraft,
  title     = {{SceneCraft: An LLM Agent for Synthesizing 3D Scenes as Blender Code}},
  author    = {Ziniu Hu and Ahmet Iscen and Aashi Jain and Thomas Kipf and Yisong Yue and David A. Ross and Cordelia Schmid and Alireza Fathi},
  booktitle = {International Conference on Machine Learning (ICML)},
  volume    = {235},
  series    = {Proceedings of Machine Learning Research},
  pages     = {19252--19282},
  year      = {2024}
}

@inproceedings{huang2024vbench,
  title     = {{VBench: Comprehensive Benchmark Suite for Video Generative Models}},
  author    = {Ziqi Huang and Yinan He and Jiashuo Yu and Fan Zhang and Chenyang Si and Yuming Jiang and Yuanhan Zhang and Tianxing Wu and Qingyang Jin and Nattapol Chanpaisit and Yaohui Wang and Xinyuan Chen and Limin Wang and Dahua Lin and Yu Qiao and Ziwei Liu},
  booktitle = {IEEE/CVF Conference on Computer Vision and Pattern Recognition (CVPR)},
  pages     = {21807--21818},
  year      = {2024}
}

@article{kong2024hunyuanvideo,
  title   = {{HunyuanVideo: A Systematic Framework For Large Video Generative Models}},
  author  = {Weijie Kong and Qi Tian and Zijian Zhang and Rox Min and Zuozhuo Dai and Jin Zhou and Jiangfeng Xiong and Xin Li and Bo Wu and Jianwei Zhang and others},
  journal = {arXiv preprint arXiv:2412.03603},
  year    = {2024}
}

@article{li2026videococo,
  title   = {{VideoCoCo: Code-as-CoT for Physically-Consistent Video Generation via an Agentic Dual-Engine System}},
  author  = {Haodong Li and Tianfei Ren and Xiaoxiao Ma and Chunmei Qing and Zhen Fang and Sipeng He and Ziyu Guo and Haoyu Wu and Juanxi Tian and Yihang Zou and others},
  journal = {arXiv preprint arXiv:2607.27380},
  year    = {2026}
}

@inproceedings{luo2026shotstream,
  title     = {{ShotStream: Streaming Multi-Shot Video Generation for Interactive Storytelling}},
  author    = {Yawen Luo and Xiaoyu Shi and Junhao Zhuang and Yutian Chen and Quande Liu and Xintao Wang and Pengfei Wan and Tianfan Xue},
  booktitle = {European Conference on Computer Vision (ECCV)},
  pages     = {193--209},
  year      = {2026},
  doi       = {10.1007/978-3-032-37261-1_12}
}

@inproceedings{meng2025holocine,
  title     = {{HoloCine: Holistic Generation of Cinematic Multi-Shot Long Video Narratives}},
  author    = {Yihao Meng and Hao Ouyang and Yue Yu and Qiuyu Wang and Wen Wang and Ka Leong Cheng and Hanlin Wang and Shuailei Ma and Yixuan Li and Cheng Chen and Yanhong Zeng and Xing Zhu and Yujun Shen and Huamin Qu},
  booktitle = {IEEE/CVF Conference on Computer Vision and Pattern Recognition (CVPR)},
  pages     = {461--471},
  year      = {2026}
}

@article{shao2024deepseekmath,
  title   = {{DeepSeekMath: Pushing the Limits of Mathematical Reasoning in Open Language Models}},
  author  = {Zhihong Shao and Peiyi Wang and Qihao Zhu and Runxin Xu and Junxiao Song and Xiao Bi and Haowei Zhang and Mingchuan Zhang and Y. K. Li and Y. Wu and Daya Guo},
  journal = {arXiv preprint arXiv:2402.03300},
  year    = {2024}
}

@inproceedings{sun20253dgpt,
  title     = {{3D-GPT: Procedural 3D Modeling with Large Language Models}},
  author    = {Chunyi Sun and Junlin Han and Weijian Deng and Xinlong Wang and Zishan Qin and Stephen Gould},
  booktitle = {International Conference on 3D Vision (3DV)},
  pages     = {1253--1263},
  year      = {2025},
  doi       = {10.1109/3DV66043.2025.00119}
}

@inproceedings{sun2026januscoder,
  title     = {{JanusCoder: Towards a Foundational Visual-Programmatic Interface for Code Intelligence}},
  author    = {Qiushi Sun and Jingyang Gong and Yang Liu and Qiaosheng Chen and Lei Li and Kai Chen and Qipeng Guo and Ben Kao and Fei Yuan},
  booktitle = {International Conference on Learning Representations (ICLR)},
  pages     = {138527--138567},
  year      = {2026}
}

@article{wan2025wan,
  title   = {{Wan: Open and Advanced Large-Scale Video Generative Models}},
  author  = {{Wan Team} and Ang Wang and Baole Ai and Bin Wen and Chaojie Mao and Chen-Wei Xie and Di Chen and Feiwu Yu and others},
  journal = {arXiv preprint arXiv:2503.20314},
  year    = {2025}
}

@inproceedings{wang2023motionctrl,
  title     = {{MotionCtrl: A Unified and Flexible Motion Controller for Video Generation}},
  author    = {Zhouxia Wang and Ziyang Yuan and Xintao Wang and Yaowei Li and Tianshui Chen and Menghan Xia and Ping Luo and Ying Shan},
  booktitle = {ACM SIGGRAPH Conference Papers},
  pages     = {1--11},
  year      = {2024},
  doi       = {10.1145/3641519.3657518}
}

@inproceedings{wang2025multishotmaster,
  title     = {{MultiShotMaster: A Controllable Multi-Shot Video Generation Framework}},
  author    = {Qinghe Wang and Xiaoyu Shi and Baolu Li and Weikang Bian and Quande Liu and Huchuan Lu and Xintao Wang and Pengfei Wan and Kun Gai and Xu Jia},
  booktitle = {IEEE/CVF Conference on Computer Vision and Pattern Recognition (CVPR)},
  pages     = {16268--16278},
  year      = {2026}
}

@article{wang2026videocoder,
  title   = {{Video-Coder: Reinforcing Agentic Coding for Video Generation}},
  author  = {Chaoyang Wang and Sicheng Gao and Junan Hu and Kunyang Li and Dongyang Chen and Yiyang Zhou and Huaxiu Yao and Yong Jae Lee and Radu Timofte and Yuzhang Shang},
  journal = {arXiv preprint arXiv:2611.09894},
  year    = {2026}
}

@inproceedings{wu2026campilot,
  title     = {{CamPilot: A Multi-Agent Cinematic Assistant for Camera-Controlled Movie Generation}},
  author    = {Yang Wu and Stefano Petrangeli and Ishita Dasgupta and Yu Shen},
  booktitle = {Proceedings of the Second Workshop for Research on Agent Language Models (REALM)},
  pages     = {1551--1564},
  year      = {2026},
  doi       = {10.18653/2026.realm-1.105}
}

@inproceedings{yang2024cogvideox,
  title     = {{CogVideoX: Text-to-Video Diffusion Models with An Expert Transformer}},
  author    = {Zhuoyi Yang and Jiayan Teng and Wendi Zheng and Ming Ding and Shiyu Huang and Jiazheng Xu and Yuanming Yang and Wenyi Hong and Xiaohan Zhang and Guanyu Feng and others},
  booktitle = {International Conference on Learning Representations (ICLR)},
  pages     = {83048--83077},
  year      = {2025}
}

@inproceedings{yang2024holodeck,
  title     = {{Holodeck: Language Guided Generation of 3D Embodied AI Environments}},
  author    = {Yue Yang and Fan-Yun Sun and Luca Weihs and Eli VanderBilt and Alvaro Herrasti and Winson Han and Jiajun Wu and Nick Haber and Ranjay Krishna and Lingjie Liu and Chris Callison-Burch and Mark Yatskar and Aniruddha Kembhavi and Christopher Clark},
  booktitle = {IEEE/CVF Conference on Computer Vision and Pattern Recognition (CVPR)},
  pages     = {16227--16237},
  year      = {2024}
}

@inproceedings{yang2026shotverse,
  title     = {{ShotVerse: Advancing Cinematic Camera Control for Text-Driven Multi-Shot Video Creation}},
  author    = {Songlin Yang and Zhe Wang and Xuyi Yang and Songchun Zhang and Xianghao Kong and Taiyi Wu and Xiaotong Zhao and Ran Zhang and Alan Zhao and Anyi Rao},
  booktitle = {SIGGRAPH Asia Conference Papers},
  note      = {arXiv:2603.11421},
  year      = {2026}
}

@article{yin2026stateflow,
  title   = {{StateFlow: Building, Evolving, and Accessing 3D World States for Previsualization}},
  author  = {Yuyang Yin and Zixiang Li and Longxuan Deng and Hongkai Li and Shifang Zhao and Junnan Liu and Weirong Huang and Mengyu Wang and Tianxiao Fu and Yikai Wang and Peng-Shuai Wang and Xiaojie Jin and Yao Zhao and Yunchao Wei},
  journal = {arXiv preprint arXiv:2608.12314},
  year    = {2026}
}

@article{zhang2026cineevolve,
  title   = {{CineEvolve: An Agentic Self-Evolution Framework for Cinematic Multi-Shot Video Generation}},
  author  = {Jingzhe Zhang and Peixuan Zhang and Genying Li and Bing Ma and Shuchen Weng and Si Li and Boxin Shi},
  journal = {arXiv preprint arXiv:2610.29590},
  year    = {2026}
}

@article{zhang2026filmflow,
  title   = {{FilmFlow: Agentic System for 3D Virtual Film Cinematography}},
  author  = {Yiming Zhang and Jiarui Guo and Haojia Wei and Yifei Liu and Yifan Yang and Hongjie Zhang and Xue Yang and Zhihang Zhong},
  journal = {arXiv preprint arXiv:2610.30753},
  year    = {2026}
}

@inproceedings{zhao2024moviedreamer,
  title     = {{MovieDreamer: Hierarchical Generation for Coherent Long Visual Sequences}},
  author    = {Canyu Zhao and Mingyu Liu and Wen Wang and Weihua Chen and Fan Wang and Hao Chen and Bo Zhang and Chunhua Shen},
  booktitle = {International Conference on Learning Representations (ICLR)},
  pages     = {50060--50090},
  year      = {2025}
}

@inproceedings{zhou2024storydiffusion,
  title     = {{StoryDiffusion: Consistent Self-Attention for Long-Range Image and Video Generation}},
  author    = {Yupeng Zhou and Daquan Zhou and Ming-Ming Cheng and Jiashi Feng and Qibin Hou},
  booktitle = {Advances in Neural Information Processing Systems (NeurIPS)},
  volume    = {37},
  pages     = {110315--110340},
  year      = {2024}
}

@inproceedings{zou2026miseenvideo,
  title     = {{Mise-en-Video: Fine-Grained 3D Control for Multi-Shot Video Generation}},
  author    = {Yingshuang Zou and Xiaoyu Shi and Yawen Luo and Ziang Li and Qinghe Wang and Mengfei Li and Chengfeng Zhao and Xintao Wang and Pengfei Wan and Kun Gai and Yuan Liu},
  booktitle = {SIGGRAPH Asia Conference Papers},
  note      = {arXiv:2610.26801},
  year      = {2026}
}
}

\end{document}